\documentclass[letter,10pt]{article}
  \usepackage[utf8]{inputenc}
  \usepackage[round]{natbib}
  \usepackage{amsmath}
  \usepackage{subfigure}
  \usepackage{booktabs} 
  \usepackage{microtype}
  \usepackage{times}  
  \usepackage{helvet} 
\usepackage{courier}  
\usepackage[hyphens]{url}  
\usepackage{graphicx} 
\usepackage{graphicx}  
\DeclareMathOperator\erf{erf}

\usepackage[margin=1in]{geometry}

\usepackage{amssymb}
\usepackage{amsthm,amsmath}
\usepackage{algorithm,algorithmic}

\title{Estimation of Accurate and Calibrated Uncertainties in Deterministic models}
\author{Enrico Camporeale$^{1}$, Algo Car\`{e}$^{2}$\\
$^{1}$University of Colorado, Boulder, CO, USA\\
$^{2}$University of Brescia, Italy}

\begin{document}

\maketitle
\begin{abstract}
 In this paper we focus on the problem of assigning uncertainties to single-point predictions generated by a deterministic model that outputs a continuous variable. This problem applies to any state-of-the-art physics or engineering models that  have a computational cost that does not readily allow to run ensembles and to estimate the uncertainty associated to single-point predictions. Essentially, we devise a method to easily transform a deterministic prediction into a probabilistic one. We show that for doing so, one has to compromise between the accuracy and the reliability (calibration) of such a probabilistic model. Hence, we introduce a cost function that encodes their trade-off. We use the Continuous Rank Probability Score to measure accuracy and we derive an analytic formula for the reliability, in the case of forecasts of continuous scalar variables expressed in terms of Gaussian distributions. The new Accuracy-Reliability cost function is then used to estimate the input-dependent variance, given a black-box mean function, by solving a two-objective optimization problem. The simple philosophy behind this strategy is that predictions based on the estimated variances should not only be accurate, but also reliable (i.e. statistical consistent with observations). Conversely, early works based on the minimization of classical cost functions, such as the negative log probability density, cannot simultaneously enforce both accuracy and reliability. We show several examples both with synthetic data, where the underlying hidden noise can accurately be recovered, and with large real-world datasets. 
\end{abstract}

\section{Introduction}
There is a growing consensus, across many fields and applications, that forecasts should have a probabilistic nature \citep{gneiting14}. This is particularly true in decision-making scenarios where cost-loss analyses are designed to take into account the uncertainties associated to a given forecast \citep{murphy77,owens14}. Unfortunately, it is often the case that well established predictive models are completeley deterministic and thus provide single-point estimates only. For example, in engineering and applied physics, models often rely on computer simulations. A typical strategy to assign confidence intervals to deterministic predictions is to perform ensemble forecasting, that is to repeat the same simulation with slightly different initial or boundary conditions \citep{gneiting05,leutbecher08}. However, this is rather expensive and it often requires a trade-off between computational cost and accuracy of the model, especially when there is a need for real-time predictions. Likewise, the most successful applications in machine learning techniques have focused on estimating target variables, with less emphasis on the estimation of the uncertainty of the prediction, even though uncertainity estimate is becoming an important topic in the machine learning community \citep{gal16}. \\
In this paper we focus on the problem of assigning uncertainties to single-point predictions, with a particular emphasis on the requirement of calibration. When dealing with a probabilistic forecast, calibration is as important as accuracy. Calibration, also known as reliability (for instance, in the meteorological literature), is the requirement that the probabilities should give an estimate of the expected frequencies of the event occurring, that is a statistical consistence between predictions and observations \citep{gneiting2007,johnson09}. 

We restrict our attention on predictive models that output a scalar continuous variable, and whose uncertainties are in general input-dependent. For the sake of simplicity, and for its widespread use, we assume that the probabilistic forecast that we want to generate is in the form of a Gaussian distribution. Hence, the problem can be cast in terms of the estimation of the input-dependent variance associated to a normal distribution centered around forecasted values provided by a model.\\
In the machine learning community, elegant and practical ways of deriving uncertainties based on non-parametric Bayesian methods are well established, either based on Bayesian neural networks \citep{mackay92,neal12,hernandez15}, deep learning \citep{gal16}, or Gaussian Processes (GPs) \citep{rasmussen06}. 
However, it is important to emphasize that whilst in the classical heteroskedastic regression problem, one is interested in learning simultaneously the mean function $f(\mathbf{x})$ and the variance $\sigma^2(\mathbf{x})$, here we assume that the mean function is provided by a black-box model (for instance, a physics simulation) that cannot easily be improved, hence the whole attention is focused on the variance estimation. This is realistic in several applied fields, where decades of work have resulted in very accurate physics-based models, that however suffer the drawback of being completely deterministic. Hence, we decouple the problem of learning mean function and variance, focusing solely on the latter. Also, it is important to keep in mind that we aim at estimating the variance using a single mean function, and not an ensemble.

\subsection{Summary of Contributions and Novelty}
The task of generating uncertainties associated with black-box predictions, thus transforming a deterministic model into a probabilistic one, and simultaneously ensuring that such uncertainties are both accurate and calibrated is novel. The closest early works in the machine learning literature that are worth mentioning are concerned with post-processing calibration. In that case, a model outputs probabilistic predictions that are not well-calibrated and the task is to re-calibrate these outputs by deriving a function $[0,1]\rightarrow [0,1]$ that maps the original probabilities to new, well-calibrated probabilities. Re-calibration has been studied extensively in the context of classification, with methods such as Platt scaling \citep{platt1999}, isotonic regression \citep{zadrozny2001}, temperature scaling \citep{guo2017}. Applications to regression is less studied. A recent work is \citet{kuleshov2018}, where isotonic regression is used to map the predicted cumulative distribution function of a continuous target variable to the observed one, effectively re-calibrating the prediction. This approach has later been criticized for not being able to distinguish between informative and non-informative uncertainty predictions \citep{levi2019} and for not being able to ensure calibration for a specific prediction (but only in an average sense) \citep{song2019}.
Finally, a relevant approach has recently been proposed in \citet{lakshminarayanan17}, building on the original idea of \citet{weigend94} of designing a neural network that outputs simultaneously mean and variance of a Gaussian distribution, by minimizing a proper score, namely the negative log likelihood of the predictive distribution. \citet{lakshminarayanan17} point out the importance of calibration of probabilistic models, even though in their work calibration is not explicitly enforced.\\
Overall, it appears that none of the previous works has recognized that calibration is only one aspect of a two-objective optimization problem. In fact, we will demonstrate that calibration (reliability) is competing with accuracy (sharpness) and therefore one must seek for the optimal trade-off between these two equally important qualities of a probabilistic forecast.\\
Our method is very general and does not depend on any particular choice for the black-box model that predicts the output targets (which indeed is not even required; all that is needed are the errors between predictions and real targets). 
The philosophy is to introduce a cost function which encodes a trade-off between the accuracy and the reliability of a probabilistic forecast. Assessing the goodness of a forecast through proper scores, such as the Negative Log Probability Density, or the Continuous Rank Probability Score, is a common practice in many applications, like weather predictions \citep{matheson76,brocker07}. Also, the notion that a probabilistic forecast should be well calibrated, or statistically consistent with observations, has been discussed at length in the atmospheric science literature \citep{murphy92,toth03}. However, the basic idea that these two metrics (accuracy and reliability) can be combined to estimate the empirical variance from a sample of observations, and possibly to reconstruct the underlying noise as a function of the inputs has never been proposed. Moreover, as we will discuss, the two metrics are competing, when interpreted as functions of the variance only. Hence, this gives rise to a two-objective optimization problem, where one is interested in achieving a good trade-off between these two properties.\\
Our main contributions are the introduction of the Reliability Score (RS), that measures the discrepancy between empirical and ideal calibration, and the Accuracy-Reliability (AR) cost function. We show that for a Gaussian distribution the RS has a simple analytical formula.
The accuracy part of the AR cost function is measured by means of the Continuous Rank Probability Score, that we argue has better properties than the more standard Negative Log Probability Density.\\
The paper is organized as follows. We first introduce the Negative Logarithm of the Probability Density and the Continuous Rank Probability Score as scores for accuracy. We then comment on the reliability, how to construct a reliability diagram for continuous probabilistic forecast, and we show that accuracy does not implies reliability and indeed the two metrics are competing. We then introduce a new score to measure reliability for Gaussian distributions and the Accuracy-Reliability score. Finally, we show how the new score can be used to estimate uncertaintiy both in toy and real-world examples.

\section{Loss functions for Accuracy}
A standard way of estimating the empirical variance of a Gaussian distribution is by maximizing its likelihood with respect to a set of observations. In practice, one can use a loss function based on the Negative Logarithm of the Probability Density (NLPD):
\begin{equation}
 {NLPD}(\varepsilon,\sigma)=\frac{\log\sigma^2}{2}+\frac{\varepsilon^2}{2\sigma^2}+\frac{\log 2\pi}{2},
\end{equation}
where we define $\varepsilon=y^o-\mu$ as the error between a given observation $y^o$ and the corresponding prediction $\mu$.
Here, we propose to use the Continuous Rank Probability Score (CRPS), in lieu of the NLPD. 
CRPS is a generalization of the well-known Brier score \citep{wilks11}, used to assess the probabilistic forecast of continuous scalar variables, when 
the forecast is given in terms of a probability density function, or its cumulative distribution. CRPS is defined as 
\begin{equation}
  {CRPS} = \int_{-\infty}^\infty \left[C(y) - H(y-y^o) \right]^2 dy
\end{equation}
where $C(y)$ is the cumulative distribution (cdf) of the forecast, $H(y)$ is the Heaviside function, and $y^o$ is the true (observed) value of the forecasted variable. For Gaussian distributions, the forecast is simply given by the mean value $\mu$
and the variance $\sigma^2$, and in this case the CRPS can be calculated analytically \citep{gneiting05} as

\begin{equation}\label{CRPS}
{CRPS}(\mu,\sigma,y^o) = \sigma\left[\frac{y^o-\mu}{\sigma}\erf\left(\frac{y^o-\mu}{\sqrt{2}\sigma}\right) + \right. 
 \left.\sqrt{\frac{2}{\pi}}\exp\left(-\frac{(y^o-\mu)^2}{2\sigma^2} \right) -\frac{1}{\sqrt{\pi}}\right]
\end{equation}

\begin{figure}[!htb]
\minipage{0.5\columnwidth}
 \center \includegraphics[width=\columnwidth]{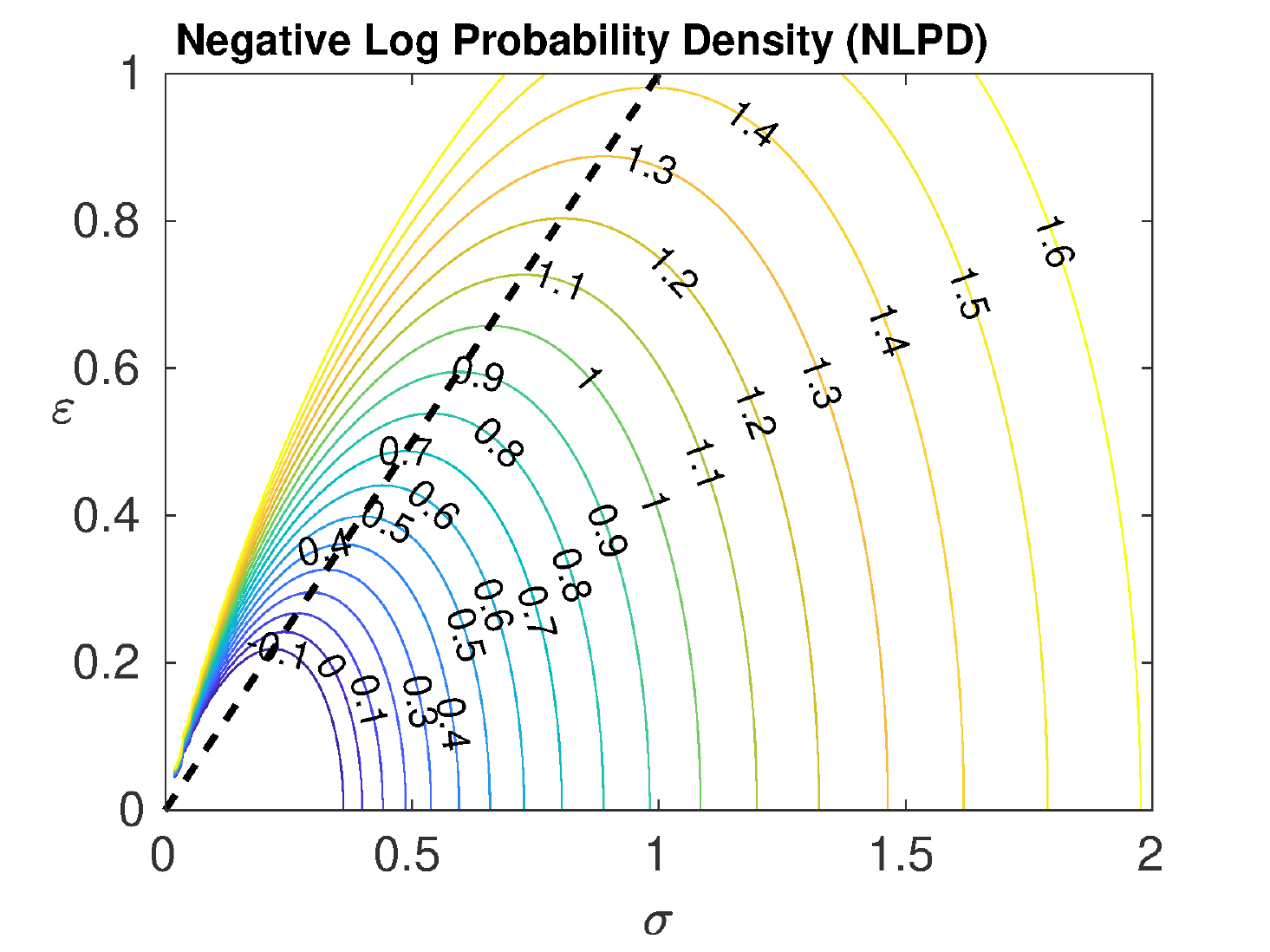}
\endminipage\hfill
\minipage{0.5\columnwidth}
 \center \includegraphics[width=\columnwidth]{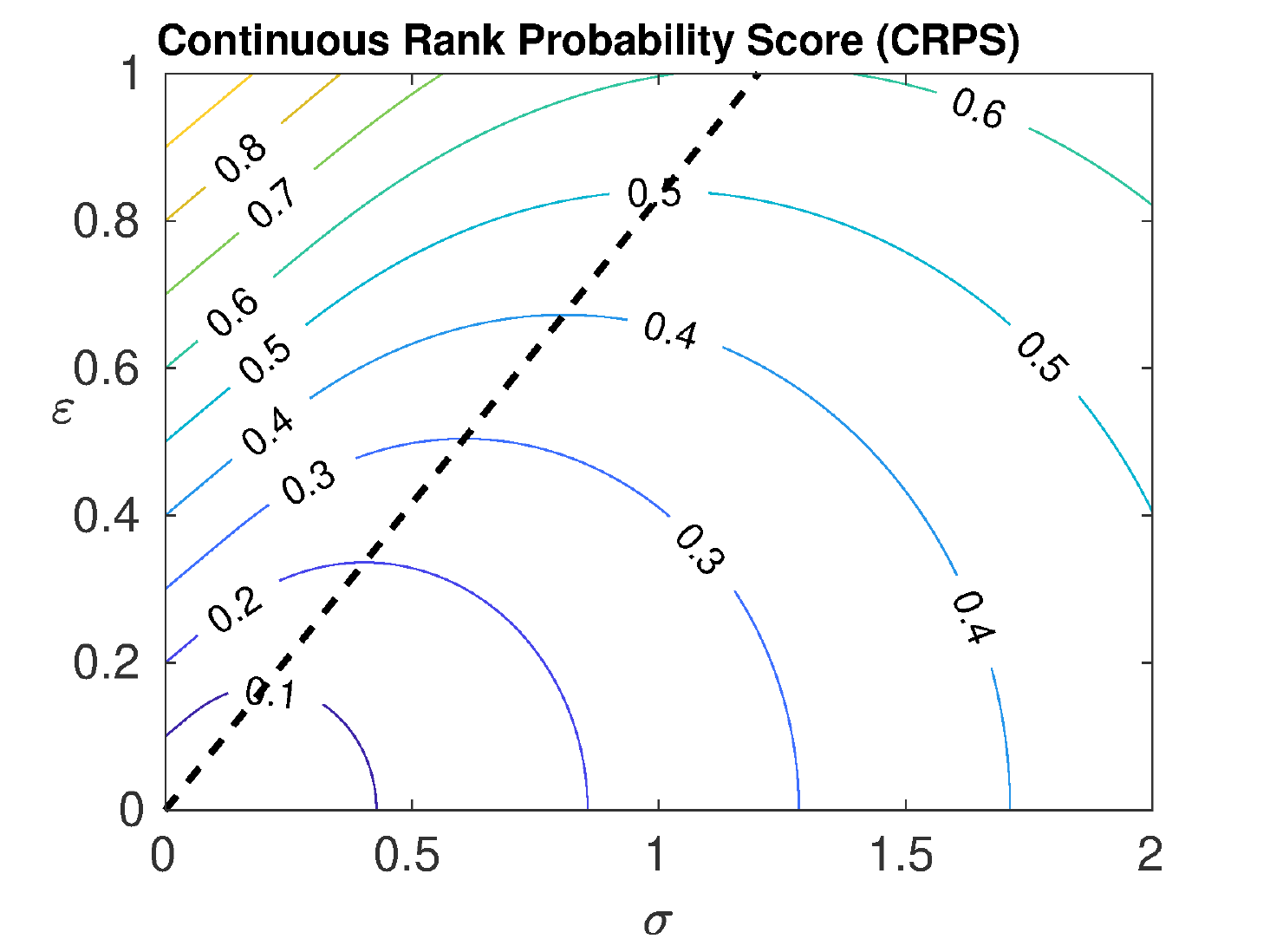}
  \endminipage
\caption{Isoline of constant NLPD (left) and CRPS (right) as a function of standard deviation $\sigma$, and error $\varepsilon$. The black dashed line indicates the minimum, as function of $\varepsilon$.}
\label{NLPD_vs_CRPS}
\end{figure}

Several interesting properties of the CRPS have been studied in the literature. Notably, its decomposition into reliability and uncertainty has been shown in \citet{hersbach00}.There are several reasons for preferring CRPS to NLPD.
They are both negatively oriented, but CRPS is equal to zero for a perfect forecast with no uncertainty (deterministic). 
Indeed, the CRPS has the same unit as the variable of interest, and it collapses to the Absolute Error $|y^o-\mu|$ for $\sigma\rightarrow 0$, that is when the forecast becomes deterministic. On the other hand, the limit $\sigma\rightarrow 0$ is problematic for NLPD (being finite only for $\varepsilon=0$).
Figure \ref{NLPD_vs_CRPS} shows a graphical comparison between NLPD (top panel) and CRPS (bottom panel). 
Different curves show the isolines for the two scores, as a function of the error $\varepsilon$ (vertical axis) and the standard deviation $\sigma$ (horizontal axis). The black dashed line indicates the minimum value of the score, for a fixed value of $\varepsilon$.
Because we are approaching the problem of variance estimation by assigning an empirical variance to single-point black-box predictions, it makes sense to minimize a score as a function of $\sigma$ only, for a fixed value of the error $\varepsilon=y^0-\mu$.
By differentiating Eq.(\ref{CRPS}) with respect to $\sigma$, one obtains
\begin{equation}
 \frac{d {CRPS}}{d\sigma} = \sqrt{\frac{2}{\pi}}\exp\left(-\frac{\varepsilon^2}{2\sigma^2} \right) -\frac{1}{\sqrt{\pi}}
\end{equation}
and the minimizer is found to be $ \sigma_{{min, CRPS}}^{2} = {\varepsilon^2}/{{\log 2}}.$
Note that the minimizer for NLPD is $\sigma_{{min, NLPD}}^{2} = \varepsilon^2$.

As it is evident from Figure \ref{NLPD_vs_CRPS}, CRPS penalizes under- and over-confident predictions in a much more symmetric way than NLPD.
Both scores are defined for a single instance of forecast and observation, hence they are  usually averaged over an ensemble of predictions, to obtain the score relative to a given model, for instance: $\overline{{CRPS}} = \sum_k {CRPS}(\mu_k,\sigma_k,y^o_k)$. 

\section{Reliability}
\begin{figure}[!htb]
\minipage{0.5\columnwidth}
 \center \includegraphics[width=\columnwidth]{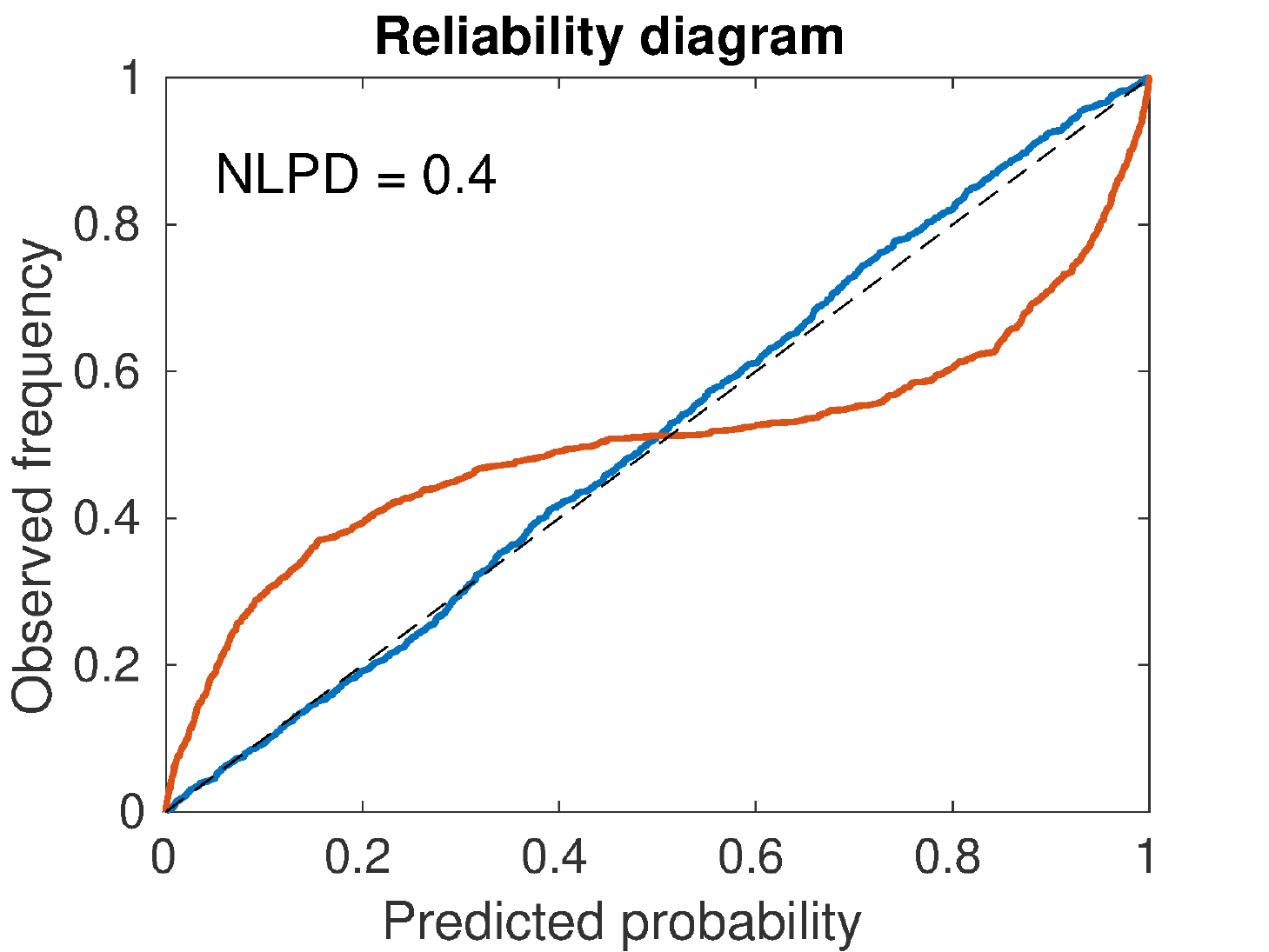}
\endminipage\hfill
\minipage{0.5\columnwidth}
 \center \includegraphics[width=\columnwidth]{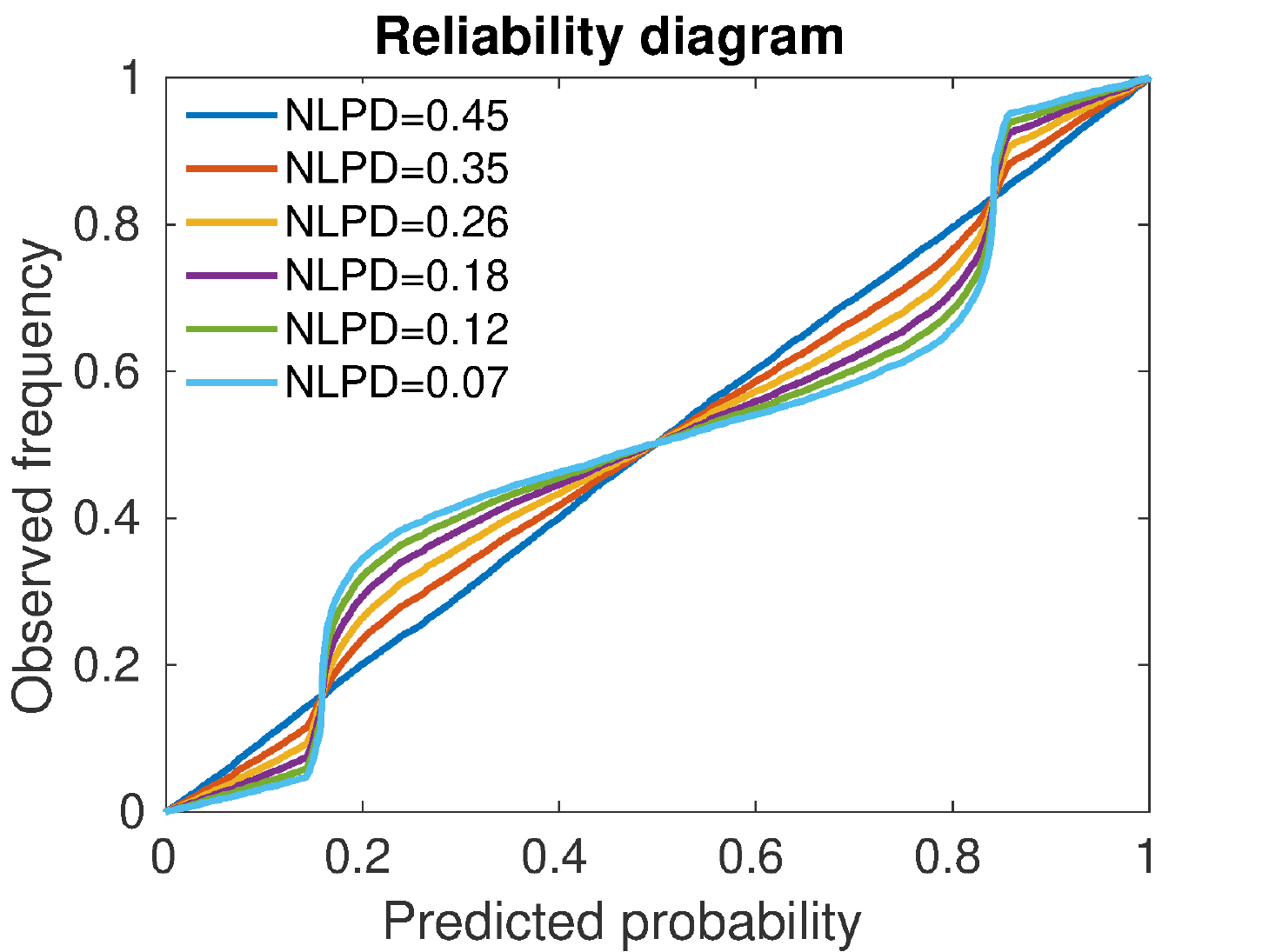}
\endminipage
\caption{Left: example of two models with identical value of NLPD=0.4, and different reliability diagram. Right: Example of several models for which the NLPD decreases (from NLPD=0.45 for the blue line to NLPD=0.07 for the cyan line), at the expense of reliability. See text for details of how the synthetic data has been generated.}
\label{fig:ex_reliability}
\end{figure}

An important consideration is that scores such as NLPD and CRPS do not automatically enforce a correct model calibration. Calibration is the property of a probabilistic model that measures its statistical consistence with observations. For forecasts of discrete events, it measures if an event predicted with probability $p$ occurs, on average, with frequency $p$. 
This concept can be extended to forecasts of a continuous scalar quantity by examining the so-called reliability diagram \citep{anderson96,hamill97,hamill01}. Note that in this paper we use the terms calibration and reliability interchangeably. A reliability diagram is produced in the following way. One collects the values of the probability predicted at all observed points, that is $P(y\leq y^o)$, which for a Gaussian distribution can be expressed analytically and we denote with $\Phi_i=\frac{1}{2}(\erf(\eta_i)+1)$, with $\eta_i=\varepsilon_i/(\sqrt{2}\sigma_i)$ being the standardized errors (the index $i$ denotes that the error is associated to the $i-$th observation/prediction in a set of size $N$). The empirical cumulative distribution of $\Phi_i$, defined as $C(y)=\frac{1}{N}\sum_{i=1}^N H(y-\Phi_i)$ ($H$ is the Heaviside function), provides the reliability diagram, with the obvious interpretation of observed frequency as a function of the  predicted probability (note that this method of producing a reliability diagram does not require binning). A perfect calibration shows in the reliability diagram as a straight diagonal line. \\
The motivating argument of this work is that two models with identical accuracy score (and we use here NLPD to illustrate the argument, but the same would be true for CRPS) can have remarkably different reliability diagrams. We show an example in Figure \ref{fig:ex_reliability}. 1000 data points have been generated as $\mathcal{N}(0,\sigma(x)^2)$, with $x\in[0,1]$ and $\sigma(x)=x+\frac{1}{2}$, as in the synthetic dataset proposed in \citet{goldberg98}. A model completely consistent with the data generation mechanism (i.e. with zero mean and variance $\sigma^2$) produces the blue line in the reliability diagram in the top panel, that is almost perfect calibration. However, one can generate a second model with a modified (wrong) variance $\tilde{\sigma}^2$ such that $ {NLPD}(\varepsilon,\tilde{\sigma})= {NLPD}(\varepsilon,\sigma)$, that is
\begin{equation}\label{ex_reliability}
\frac{\log\tilde{\sigma}^2}{2}+\frac{\varepsilon^2}{2\tilde{\sigma}^2} =\frac{\log\sigma^2}{2}+\frac{\varepsilon^2}{2\sigma^2}
\end{equation}
Eq. (\ref{ex_reliability}) always produces a solution $\tilde{\sigma}\neq\sigma$, as long as $\sigma^2\neq\varepsilon^2$ (the global minimum of NLPD, for fixed $\varepsilon$). Graphically this can be seen in Figure \ref{NLPD_vs_CRPS}: for a constant $\varepsilon$ value, there are two values of $\sigma$ on the same NLPD contour. The red line in the top panel of Figure \ref{fig:ex_reliability} has been derived from such a modified model $\mathcal{N}(0,\tilde{\sigma}^2)$, which is obviously mis-calibrated. For this example NLPD=0.4 (equal for both cases). As a complementary argument, we show in the bottom panel of Figure \ref{fig:ex_reliability} the reliability diagram of several models, with decreasing values of NLPD. One can appreciate that progressively decreasing NLPD results in a worse and worse calibration (note that NLPD is negatively oriented). These models have been generated again starting from the perfectly calibrated synthetic model, progressively shifting the values assigned to $\sigma^2$, towards the global minimum $\sigma^2=\varepsilon^2$ (hence decreasing NLPD). Thus, minimizing a traditional cost function such as NLPD does not necessarily implies to achieve a well-calibrated model. Of course, we are not suggesting that any model generated by means of minimizing NLPD is inevitably mis-calibrated. However, unless explicitly enforced, calibration will be a by-product of other properties. Once again, the same is true for CRPS.

\subsection{Reliability Score for Gaussian forecast}
Reliability is a statistical property of a model, defined for a large enough ensemble of forecasts-observations. Here, we introduce the reliability score for normally distributed forecasts. In this case, we expect the standardized errors $\eta$ calculated over a sample of $N$ predictions-observations to have a standard normal distribution with cdf $\Phi(\eta)=\frac{1}{2}(\erf(\eta)+1)$. Hence we define the Reliability Score (RS) as:
\begin{equation}\label{RS_1}
 {RS} = \int_{-\infty}^\infty \left[\Phi(\eta) - C(\eta)\right]^2 d\eta
\end{equation}
where $C(\eta)$ is the empirical cumulative distribution of the standardized errors $\eta$, that is 
\begin{equation}
 C(y) = \frac{1}{N}\sum_{i=1}^N H(y-\eta_i)
\end{equation}
with $\eta_i = (y^o_i-\mu_i)/(\sqrt{2}\sigma_i)$. Note that each error $(y^o_i-\mu_i)$ is standardized with respect to a different (input-dependent) $\sigma_i$. RS measures the divergence of the empirical distribution of standardized errors $\eta$ from a standard normal distribution. 
From now on we will use the convention that the set $\{\eta_1,\eta_2,\ldots \eta_N\}$ is sorted ($\eta_i\leq\eta_{i+1}$). Obviously this does not imply that $\mu_i$ or $\sigma_i$ are sorted as well.
Interestingly, the integral in Eq. (\ref{RS_1}) can be calculated analytically, via expansion into a telescopic series, yielding:
\begin{equation}
 {RS} = \sum_{i=1}^N \left[\frac{\eta_i}{N}\left(\erf(\eta_i)+1\right) - \frac{\eta_i}{N^2}(2i-1) + \frac{\exp(-\eta_i^2)}{\sqrt{\pi}N}\right] -\frac{1}{2}\sqrt{\frac{2}{\pi}}\label{RS}
\end{equation}

\normalsize
Differentiating the $i$-th term of the above summation, RS$_i$, with respect to $\sigma_i$ (for fixed $\varepsilon_i$), one obtains
\begin{equation}
 \frac{d{RS}_i}{d\sigma_i} =   \frac{\eta_i}{N\sigma_i}\left(\frac{2i-1}{N}-\erf(\eta_i)-1 \right)
\end{equation}
Hence, ${RS}_i$ is minimized when the values $\sigma_{{min}}^{{RS}}$ satisfy
\begin{equation}\label{optimal_eta}
 \erf(\eta_i)=\erf\left(\frac{\varepsilon_i}{\sqrt{2}\sigma_{{min}}^{{RS}}}\right) = \frac{2i-1}{N}-1
\end{equation}
This could have been trivially derived by realizing that the distribution of $\eta_i$ that minimizes RS is the one such that 
the values $\Phi(\eta_i)$ are uniform in the interval $[0,1]$.

\section{The Accuracy-Reliability cost function}
The Accuracy-Reliability (AR) cost function introduced here follows from the simple principle that the variances $\sigma_i^2$ estimated from an ensemble of errors $\varepsilon_i$ should result in a model that is both accurate (with respect to the CRPS score), and reliable (with respect to the RS score). Clearly, this gives rise to a two-objective optimization problem. It is trivial to verify that CRPS and RS cannot simultaneously attain their minimum value (as was evident from Figure \ref{fig:ex_reliability}). Indeed, by minimizing the former, $\eta_i = \frac{1}{2}\sqrt{\log 4}$ for any $i$. On the other hand, a constant $\eta_i$ cannot result in a minimum for RS, according to Eq. (\ref{optimal_eta}). This demonstrates that methods that focus solely on re-calibration (any method of choice will have an equivalent into minimizing RS) can possibly result in the deterioration of accuracy. In passing, we note that any cost function that is minimized (for constant $\varepsilon$) by a value of the variance $\sigma^2$ that is linear in $\varepsilon^2$ suffers this problem (because $\eta_i$ will be a constant). 
Finally, notice that trying to minimize RS as a function of $\sigma_i$ (for fixed errors $\varepsilon_i$) results in an ill-posed problem, because RS is solely expressed in terms of the standardized errors $\eta$. Hence, there is no unique solution for the variances that minimizes RS. Hence, RS can be more appropriately thought of as a regularization term in the Accuracy-Reliability cost function. The simplest strategy to deal with multi-objective optimization problems is to scalarize the cost function, which we define here as
\begin{equation}\label{AR}
 {AR} = \beta\cdot \overline{{CRPS}} + (1-\beta){RS}.
\end{equation}
We choose the scaling factor $\beta$ as
\begin{equation}\label{beta}
 \beta={{RS}}_{min}/(\overline{{CRPS}}_{min} + {RS}_{min}).
\end{equation}
The minimum of $\overline{{CRPS}}$ is $\overline{{CRPS}}_{min}=\frac{\sqrt{\log 4}}{2N}\sum_{i=1}^N \varepsilon_i$, which is simply the mean of the errors, rescaled by a constant. 
The minimum of RS follows from Eqs. (\ref{RS}) and (\ref{optimal_eta}): 
\begin{equation}
{RS}_{min} = \frac{1}{\sqrt{\pi} N}\sum_{i=1}^N \exp\left(-\left[\erf^{-1}\left(\frac{2i-1}{N}-1\right)\right]^2\right)-\frac{1}{2}\sqrt{\frac{2}{\pi}}
\end{equation}
Notice that ${RS}_{min}$ is only a function of the size of the sample $N$, and it converges to zero for $N\rightarrow \infty$. 
The heuristic choice in Eq. (\ref{beta}) is justified by the fact that the two scores might have different orders of magnitude, and therefore we rescale them in such a way that they are comparable in our cost function (\ref{AR}). We believe this to be a sensible choice, although there might be applications where one would like to weigh the two scores differently.  
In future work, we will explore the possibility of optimizing $\beta$ in a principled way, for instance constraining the difference between empirical and ideal reliability score to be within limits given by the dataset size $N$, or by making $\beta$ a learnable parameter. Finally, in our practical implementation, we neglect the last constant term in the definition (\ref{RS}) so that, for sufficiently large $N$, ${RS}_{min}\simeq \frac{1}{2}\sqrt{\frac{2}{\pi}}\simeq 0.4$

\section{Results}
In summary, we want to estimate the input-dependent values of the empirical variances $\sigma_i^2$ associated to a sample of $N$ observations for which we know the errors $\varepsilon_i$. We do so by solving an optimization problem in which the set of estimated $\sigma_i$ minimizes the AR cost function defined in Eq. (\ref{AR}). This newly introduced cost function has a straightforward interpretation as the trade-off between accuracy and reliability, which are two essential but conflicting properties of probabilistic models. In practice, because we want to generate a model that is able to predict $\sigma^2$ as a function of the inputs $\mathbf{x}$ on any point of a domain, we introduce a structure that enforces a certain degree of smoothness of the unknown variance, in the form of a regression model. 
In the following we show some experiments on toy problems and on multidimensional real dataset to demonstrate the easiness, robustness and accuracy of the method.\\

\subsection{Toy problems}
In order to facilitate comparison with previous works, we choose some of the datasets used in \citet{kersting07}, although for simplicity of implementation we rescale the standard deviation so to be always smaller or equal to 1. Since in our method we assume that a mean function is provided, for the topy problems we use the result of a standard (homoskedastic) Gaussian Process regression as $f(x)$.

\begin{figure}[ht]
\minipage{0.5\columnwidth}
 \center \includegraphics[width=\columnwidth]{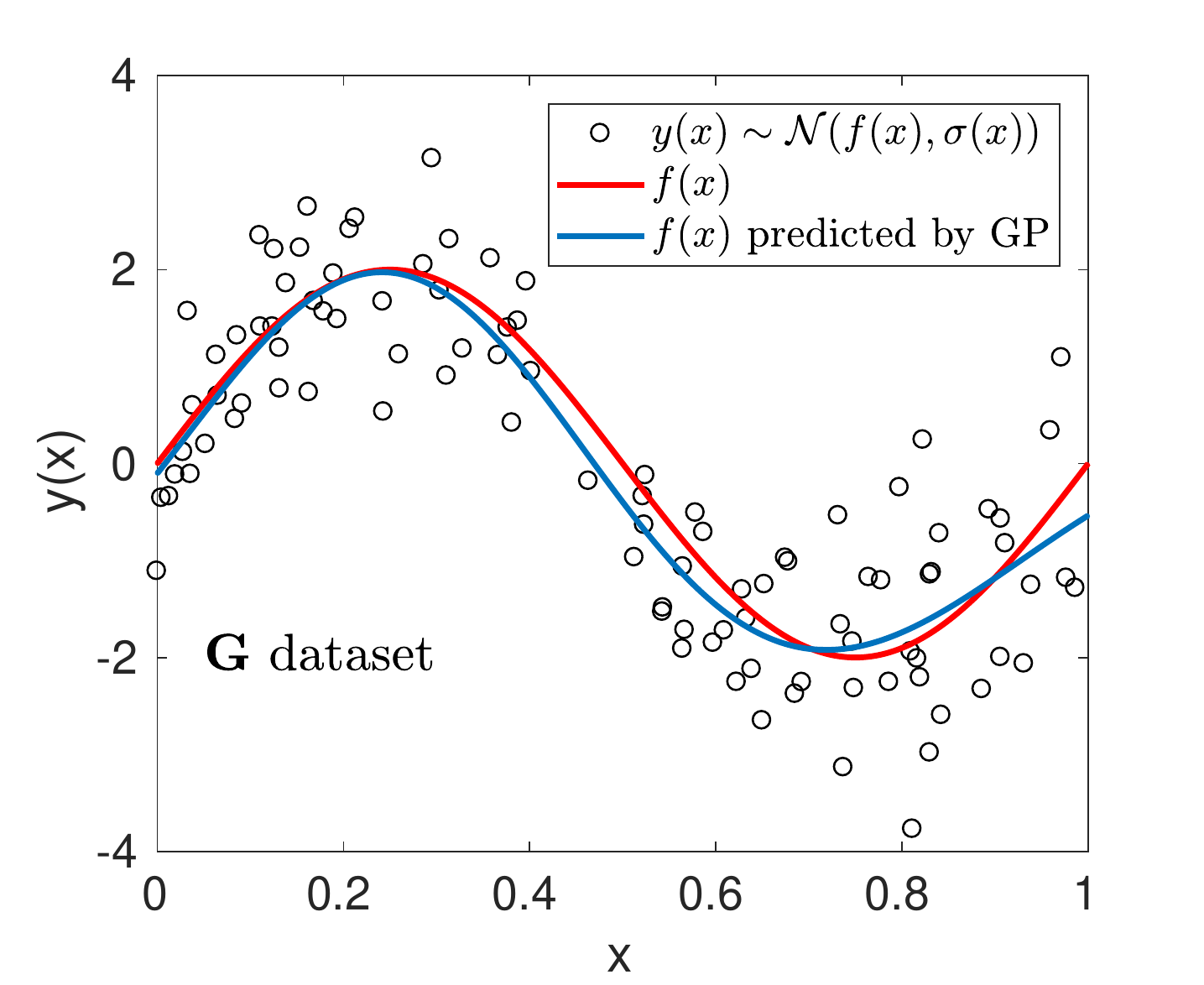}
 \center \includegraphics[width=\columnwidth]{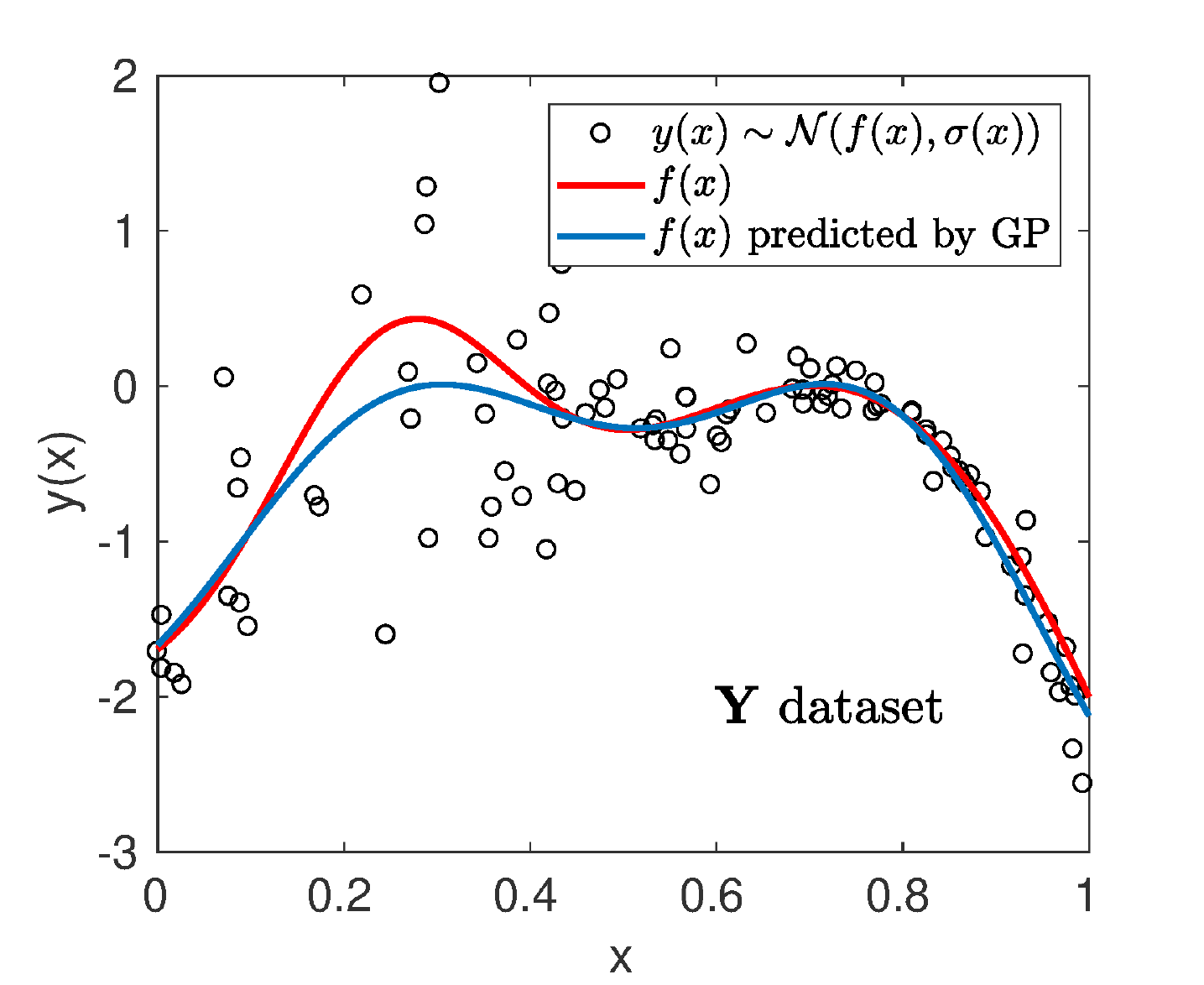}
\endminipage\hfill
 \minipage{0.5\columnwidth}
 \center \includegraphics[width=\columnwidth]{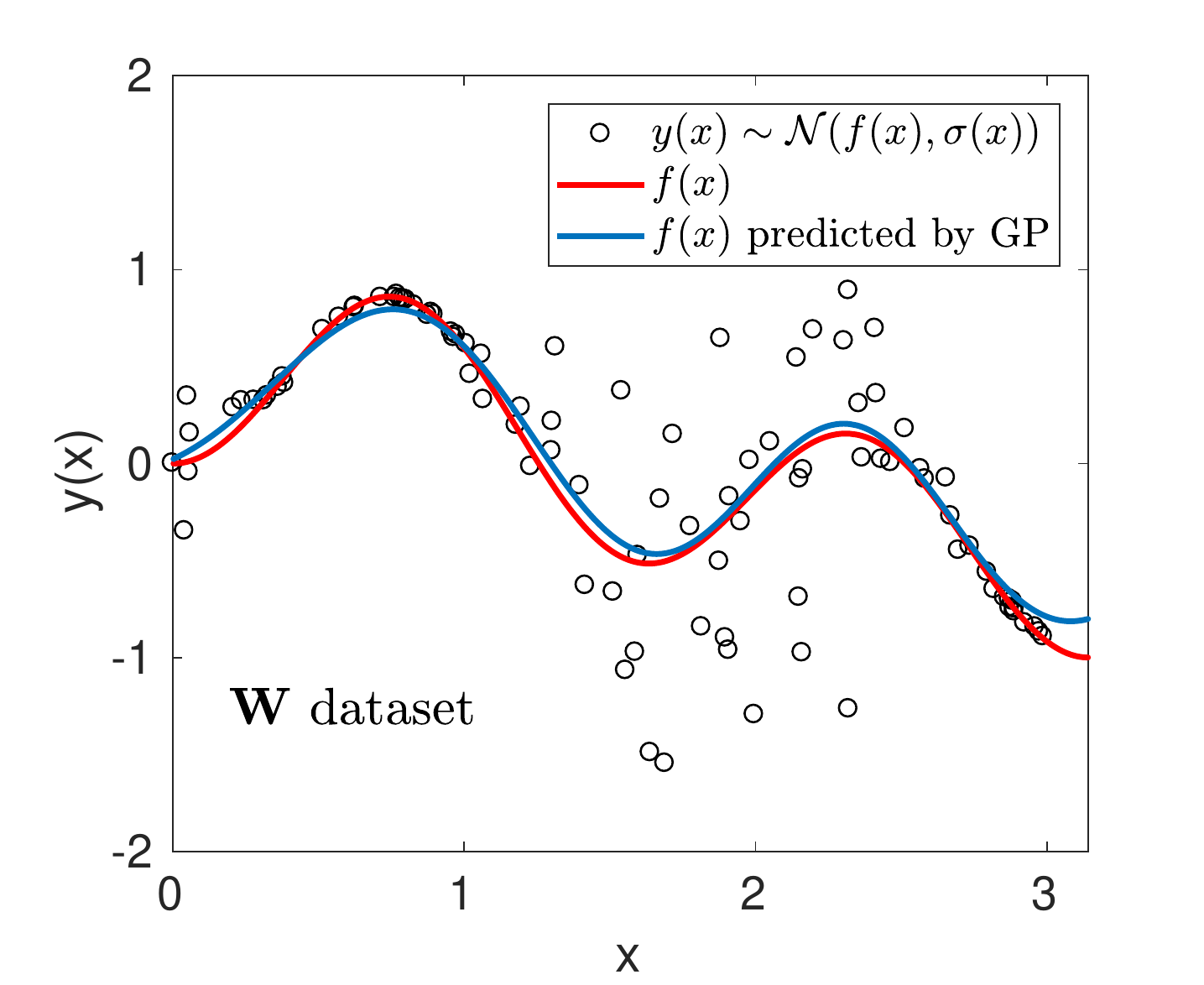}
 \center \includegraphics[width=\columnwidth]{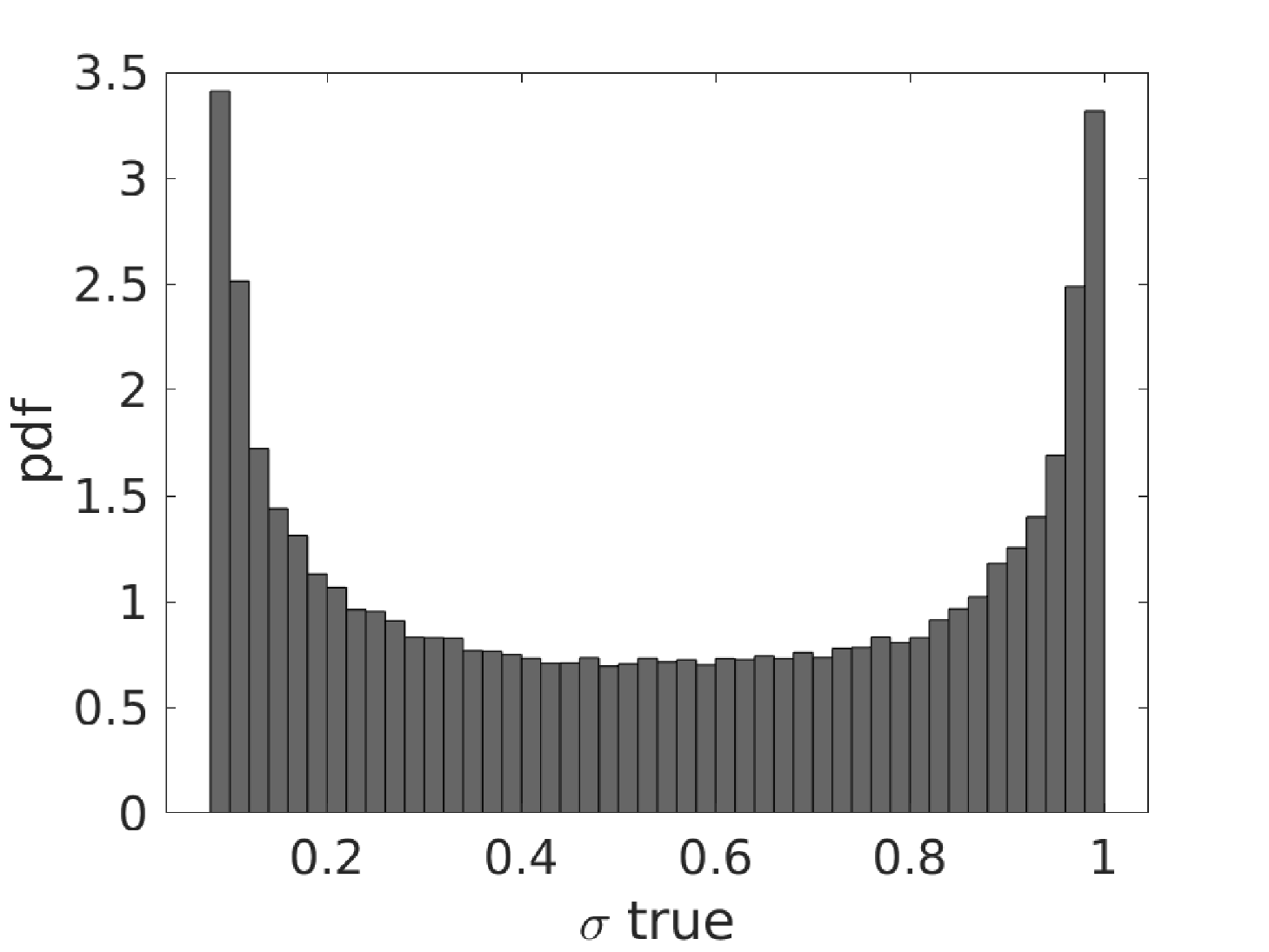}
\endminipage
\caption{Top:100 points sampled from the {\bf G}(left) and {\bf Y}(right) dataset (circles). Bottom-left: 100 points sampled from the {\bf W} dataset (circles). The blue line shows the true mean function $f(x)$, while the red one is the one predicted by the GP model. Bottom-right: Distribution of true values of standard deviation $\sigma$ for the {\bf 5D} dataset.} \label{fig:toy_regression}
\end{figure}

For all datasets the targets $y_i$ are sampled from a Gaussian distribution $\mathcal{N}(f(x),\sigma(x)^2)$. The first three datasets are one-dimensional in $x$, while in the fourth we will test the method on a five-dimensional space, thus showing the robustness of the proposed strategy.\\
{\bf G} dataset: $x \in [0,1]$, $f(x) = 2\sin(2\pi x)$, $\sigma(x) = \frac{1}{2}x+\frac{1}{2}$ \cite{goldberg98}. \\
{\bf Y} dataset: $x \in [0,1]$, $f(x) = 2(\exp(-30(x-0.25)^2)+\sin(\pi x^2))-2$, $\sigma(x) = \exp(\sin(2\pi x))/3$ \cite{yuan04}. \\
{\bf W} dataset: $x \in [0,\pi]$, $f(x) = \sin(2.5x)\sin(1.5x)$, $\sigma(x) = 0.01+0.25(1-\sin(2.5x))^2$ \cite{weigend94, williams96}. \\
{\bf 5D} dataset: $\mathbf{x} \in [0, 1]^5$, $f(\mathbf{x})=0$, $\sigma(\mathbf{x})=0.45(\cos(\pi + \sum_{i=1}^5 5x_i) + 1.2)$ \cite{genz84}.
Examples of 100 points sampled from the {\bf G, Y, W} dataset ar shown in Figure \ref{fig:toy_regression} (circles), along with the true mean function $f(x)$ (red), and the one predicted by a standard Gaussian Process regression model (blue). The bottom-right plot in Figure \ref{fig:toy_regression} shows the distribution of $\sigma$, which ranges in the interval $[0.09,0.99]$.\\
For the {\bf G}, {\bf Y}, and {\bf W} datasets the model is trained on 100 points uniformly sampled in the domain. The {\bf 5D} dataset is obviously more challenging, hence we use 10,000 points to train the model (note that this results in less points per dimension, compared to the one-dimensional tests).
For all experiments we test 100 independent runs.

We have tested a neural network and a polynomial best fit as regression model.
For simplicity, we choose a single neural network architecture, that we use for all the tests. We use a network with 2 hidden layers, respectively with 50 and 10 neurons. The activation functions are rectified linear (ReLU) and a symmetric saturating linear function, respectively. The output is given in terms of $\log\sigma$, to enforce positivity of $\sigma^2$. For all experiments, the datasets are randomly divided into training ($33\%$), validation ($33\%$) and test ($34\%$) sets. All the reported metrics are calculated on the test set only. The network is trained using a standard BFGQ quasi-Newton algorithm, and the iterations are forcefully stopped when the loss function does not decrease for 10 successive iterations on the validation set. The only inputs needed are the inputs $\mathbf{x}_i$ and the corresponding errors $\varepsilon_i$. Finally, in order to avoid local minima due to the random initialization of the neural network weights, we train five independent networks and choose the one that yields the smallest cost function.\\
In the case of low-dimensional data one might want to try simpler and faster approaches than a neural network, especially if smoothness of the underlying function $\sigma(x)$ can be assumed. For the one-dimensional test cases ({\bf G, Y, W}) we have devised a simple polynomial best fit strategy. 
We assume that $\sigma(x)$ can be approximated by a polynomial of unknown order, equal or smaller than 10: $\sigma(x)=\sum_{l=0}^{10} \theta_l x^l$, where in principle one or more $\theta_l$ can be equal to zero. The vector $\Theta=\{\theta_0,\theta_1,\ldots,\theta_{10}\}$ is initialized with $\theta_0=const$ and all the others equal to zero. The constant can be chosen, for instance, as the standard deviation of the errors $\varepsilon$. The polynomial best fit is found by means of an iterative procedure (Algorithm 1).

\begin{algorithm}[ht]
   \caption{Polynomial best fit}
   \label{alg:poly}
\begin{algorithmic}
   \STATE {\bfseries Input:} data $x_i, \varepsilon_i$
   \STATE Initialize $p = 0$, $\theta_0=const$, $P_{max}=10$, $tol$
   \WHILE{$p\leq P_{max}$ \& $err>tol$}
     \STATE ${p=p+1}$
     \STATE Initial guess for optimization $\Theta=\{\theta_0,\ldots,\theta_{p-1},0\}$
     \STATE $\Theta=\text{argmin AR}(\sigma_i)$ (with $\sigma_i = \sum_{l=0}^p \theta_lx_i^l$) 
     \STATE err = $||\text{AR}(\sigma(\Theta_{old})) - \text{AR}(\sigma(\Theta_{new}))||_2$
   \ENDWHILE 
\end{algorithmic}
\end{algorithm}
In words, the algorithm finds the values of $\Theta$ for a given polynomial order that minimizes the Accuracy-Reliability cost function. Then it tests the next higher order, by using the previous solution as initial guess. Whenever the difference between the solutions obtained with two successive orders is below a certain tolerance, the algorithm stops. The multidimensional optimization problem is solved by a BFGQ Quasi-Newton method with a cubic line search procedure. Note that whenever a given solution is found to yield a local minimum for the next polynomial order, the iterations are terminated.

The results for the 1D datasets $\bf G, Y, W$ are shown in Figures \ref{G_dataset} - \ref{W_dataset}, in a way consistent with \citet{kersting07}. The red lines denote the true standard deviation $\sigma(x)$ used to generate the data. The black line indicates the values of the estimated $\sigma$ averaged over 100 independent runs, and the gray areas represent one and two standard deviations from the mean. A certain spread in the results is due to different training sets (in each run 100 points are sampled independently) and, for the Neural Network, to random initialization.
The top panels show the results obtained with the Neural Network, while the bottom panels show the result obtained with the polynomial fit. In all cases, except for the {\bf W} dataset (polynomial case, bottom panel), the results are very accurate. 

\begin{figure}[t]
\begin{center}
{\includegraphics[width=.5\columnwidth]{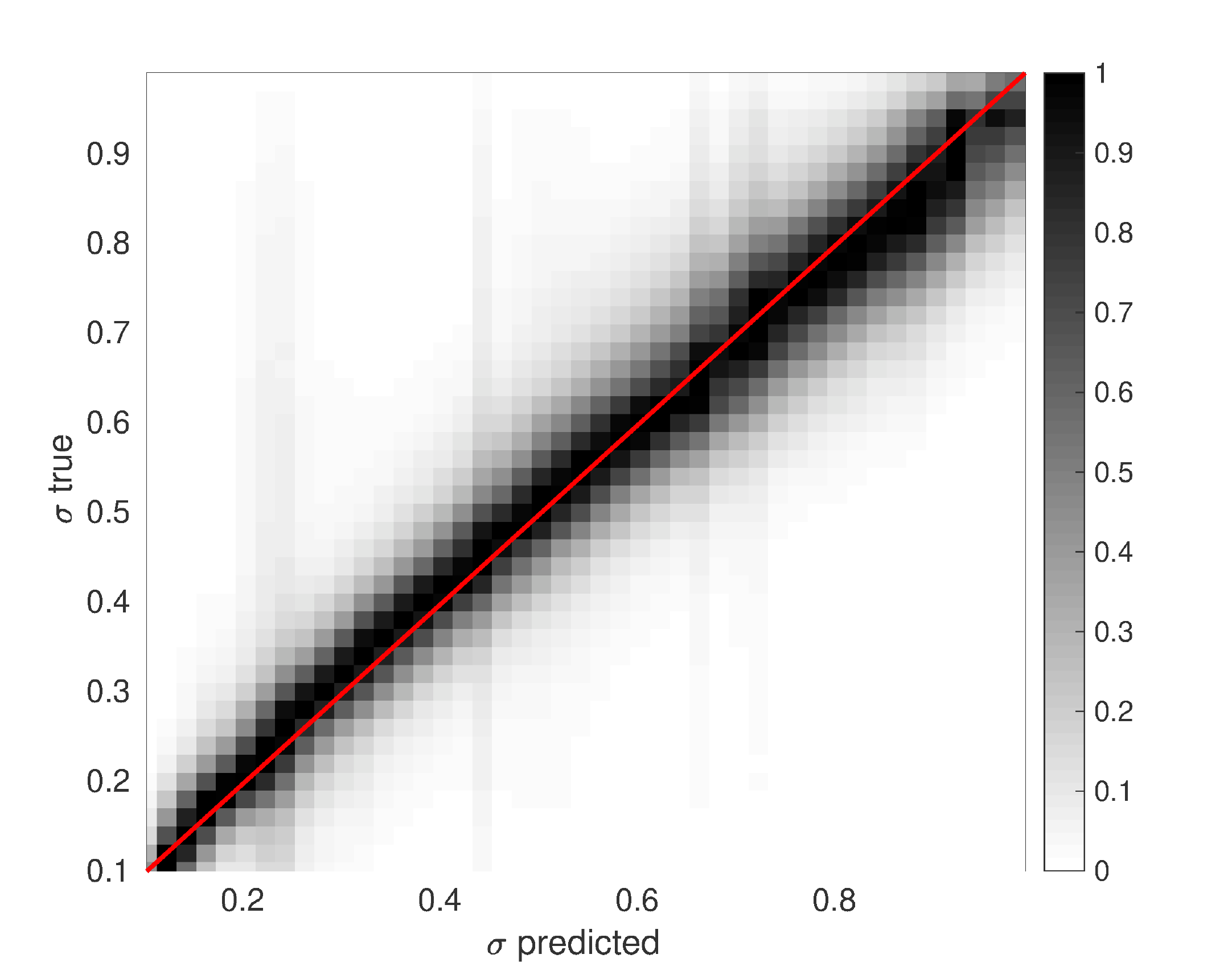}}
\caption{Probability density of the prediction versus real values of $\sigma$ for the {\bf 5D} dataset. The red line denotes perfect prediction. The densities are normalized to have maximum value along each column equal to one. 10,000,000 samples have been used to generate the plot (with a training set of 10,000 points).}
\label{multiD_1}
\end{center}
\end{figure}

For the {\bf 5D} dataset it is impractical to compare graphically the real and estimated $\sigma(\mathbf{x})$ in the 5-dimensional domain. Instead, in Figure \ref{multiD_1} we show the probability density of the real versus predicted values of the standard deviation. Values are normalized such that the maximum value in the colormap for any value of predicted $\sigma$ is equal to one (i.e. along vertical lines). The red line shows a perfect prediction. The colormap has been generated by 10e6 points, while the model has been trained with 10,000 points only.
For this case, we have used an exact mean function (equal to zero), in order to focus exclusively on the estimation of the variance.
We believe that this is an excellent result for a very challenging task, given the sparsity of the training set, that shows the robustness of the method.

\subsection{Real-World dataset}
We have tested our method on the same datasets used in \citep{hernandez15}. The only difference with the topy problems is that we use 70\% of the data for training, and we only use a neural network as regressor.
The results reported in Table \ref{table:realdata} are computed over 50 independent runs. For each run, we first train a standard neural network to provide the mean function $f(\mathbf{x})$, by minimizing the mean square errors with respect to the targets. We then compare our method against three different models (first row in Table 1): CRPS means that the variance is estimated by minimizing CRPS only; KM denotes a K-means method; RECAL indicates the recalibration method of \citep{kuleshov2018}, and AR denotes our method. 
The scores reported (second row) are the median values (calculated on the test set only) of CRPS and of the calibration error. To estimate the latter we derive the reliability diagram (in the way described in section 2), and we compute the maximum distance to the optimal reliability (straight diagonal line). This is denoted, in Table \ref{table:realdata}, as Cal. err. (in percentage). 
For the K-means method (which is possibly the simplest baseline method) we have clustered the training data in $k$ groups, calculated the standard deviation $\sigma$ for each cluster, and assigned the same value of $\sigma$ for all test points belonging to a given cluster. We have run experiments with $k$ ranging from 1 to 10, and we report the minimum values obtained for CRPS, and Cal. err. for the model that yields the best calibration. The RECAL method takes the $\sigma$ estimated by the AR method and applies the recalibration algorithm of \citep{kuleshov2018}. The training sets used for all methods are the same.
The results obtained by using the AR cost function are always better calibrated than minimizing CRPS only and than the KM method. In two cases only (Protein and Wine datasets) RECAL yields a slightly better calibration error. However, in both cases, the accuracy (CRPS) of teh RECAL method is penalized and we believe that the best trade-off is still acheived by AR.
In fact, AR offers the best trade-off between accuracy and calibration across all dataset, as expected.

\begin{table*}[ht]
\caption{Comparison between different methods on several multidimensional dataset. Median values are reported, calculated over 50 runs. Best values are in bold.}\label{table:realdata}
\centering
\begin{scriptsize}
\begin{tabular}{|ccc|cccc|cccc|}
\toprule
\multicolumn{3}{ |c| }{Method } & CRPS &RECAL& KM& AR & CRPS& RECAL& KM &AR\\
\multicolumn{3}{ |c| }{Score} &  \multicolumn{4}{ |c| }{CRPS}  &  \multicolumn{4}{ |c| }{Cal. err. ($\%$)} \\
\midrule
Dataset & Size & Dim. & \multicolumn{8}{ |c| }{}  \\
\midrule
Boston Housing  &506&13& 0.25  &   0.25  &   0.25  &   {\bf 0.23} & 26.2   &   20.6   &   17.5   &    {\bf 16.7}\\
Concrete & 1,030& 8 &  0.22  &   0.23  &   0.26  &   {\bf 0.21} & 22.6  &    14.4  &    22.1  &    {\bf 11.5}\\
Energy & 768 & 8 & 0.059  &  0.056  &  0.087 &   {\bf 0.052} & 29.3  &    29.2  &    28.3  &    {\bf 13.0}\\
Kin8nm & 8,192 & 8 & 0.17  &   {\bf 0.16}  &   0.24  &   {\bf 0.16} & 15.9   &    8.3  &    25.5  &    {\bf 5.8}\\
Power plant &  9,568 & 4 & 0.13  &   0.13  &   0.15  &   {\bf 0.12} & 12.5   &   3.4   &   16.1   &   {\bf 2.6}\\
Protein & 45,730 & 9 & 0.38  &   0.47  &   0.40  &   {\bf 0.37} & 13.1  &    {\bf 5.0}  &    10.6  &    5.4\\
Wine & 1,599 & 11 & 0.48  &   0.50  &   {\bf 0.46}  &   0.48 & 16.0  &    {\bf 7.9}  &    {8.0}  &    8.3\\
Yacht & 308 & 6 & {\bf 0.06}   &  {\bf 0.06}  &   0.19  &  {\bf 0.06} & 26.0  &    24.3  &    36.6   &   {\bf 19.5}\\

\bottomrule
\end{tabular}
\end{scriptsize}
\end{table*}

\section{Discussion and future work}
We have presented a simple parametric model for estimating the input-dependent variance of probabilistic forecasts.
We assume that the data is distributed as $\mathcal{N}(f(\mathbf{x}),\sigma(\mathbf{x})^2)$, and that an approximation of the mean function $f(\mathbf{x})$ is available (the details of the model that approximates the mean function are not important). In order to generate the variance $\sigma^2(\mathbf{x})$, we propose to minimize the Accuracy-Reliability (AR) cost function, which depends only on $\sigma$, on the errors $\varepsilon$, and on the size of the training set $N$. We have shown that the classical method of minimizing the Negative Log Probability Density (NLPD) does not guarantees that the result will be well-calibrated. On the other hand, methods that exclusively focus on the post-process calibration tend to spoil their accuracy. Indeed, we have discussed how accuracy and reliability are two conflicting metrics for a probabilistic forecast and how the latter can serve as a regularization term for the former. We have shown that by using the new AR cost function, one is able to accurately discover the hidden noise function. Several tests for synthetic and real-world (large) datasets have been shown.

\begin{figure}[h]
\vskip 0.1in
\begin{center}
\centerline{\includegraphics[width=.5\columnwidth]{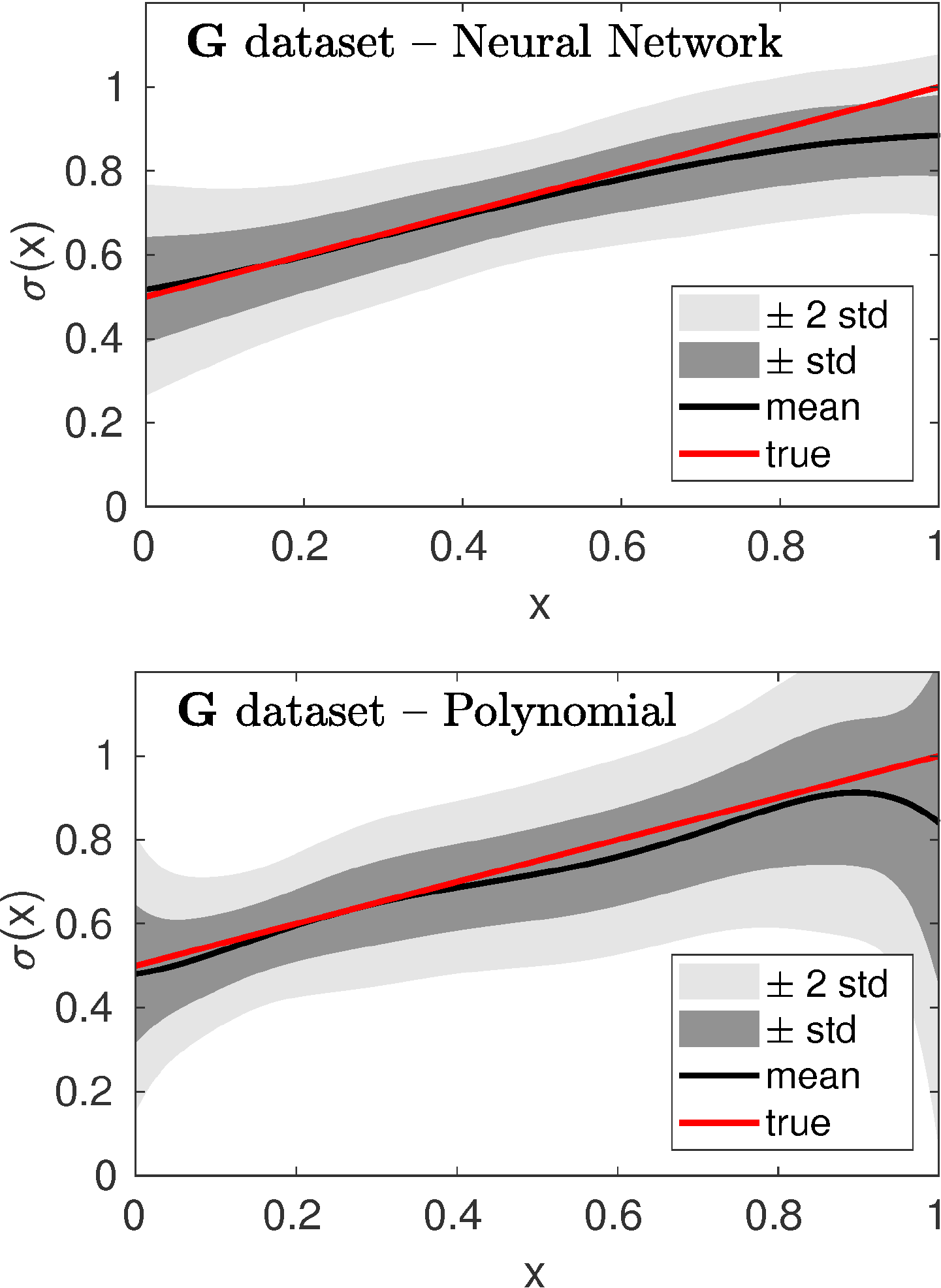}}
\caption{{\bf G} dataset: True value of the standard deviation $\sigma$ (red line) and mean value obtained averaging over 100 independent runs (black line). The gray shaded areas denote the confidence interval of one and two standard deviations calculated from the same ensemble of runs. In the top panel $\sigma$ is calculated through a Neural Network, while in the bottom panel as a polynomial function (see text). }
\label{G_dataset}
\end{center}
\vskip -0.1in
\end{figure}

\begin{figure}[ht]
\vskip 0.1in
\begin{center}
\centerline{\includegraphics[width=.5\columnwidth]{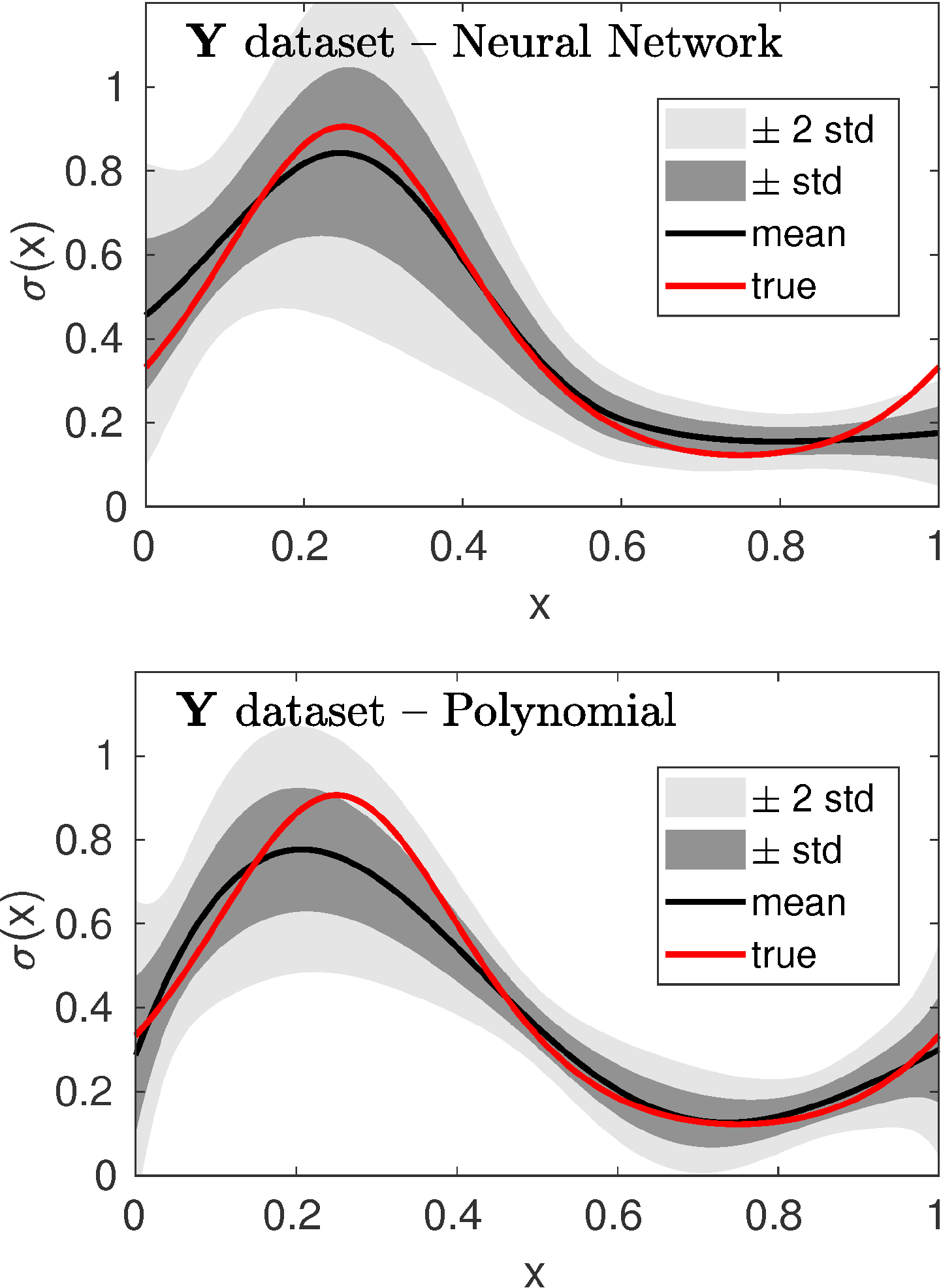}}
\caption{{\bf Y} dataset: True value of the standard deviation $\sigma$ (red line) and mean value obtained averaging over 100 independent runs (black line). The gray shaded areas denote the confidence interval of one and two standard deviations calculated from the same ensemble of runs. In the top panel $\sigma$ is calculated through a Neural Network, while in the bottom panel as a polynomial function (see text). }
x\label{Y_dataset}
\end{center}
\vskip -0.1in
\end{figure}

\begin{figure}[!ht]
\vskip 0.1in
\begin{center}
\centerline{\includegraphics[width=.5\columnwidth]{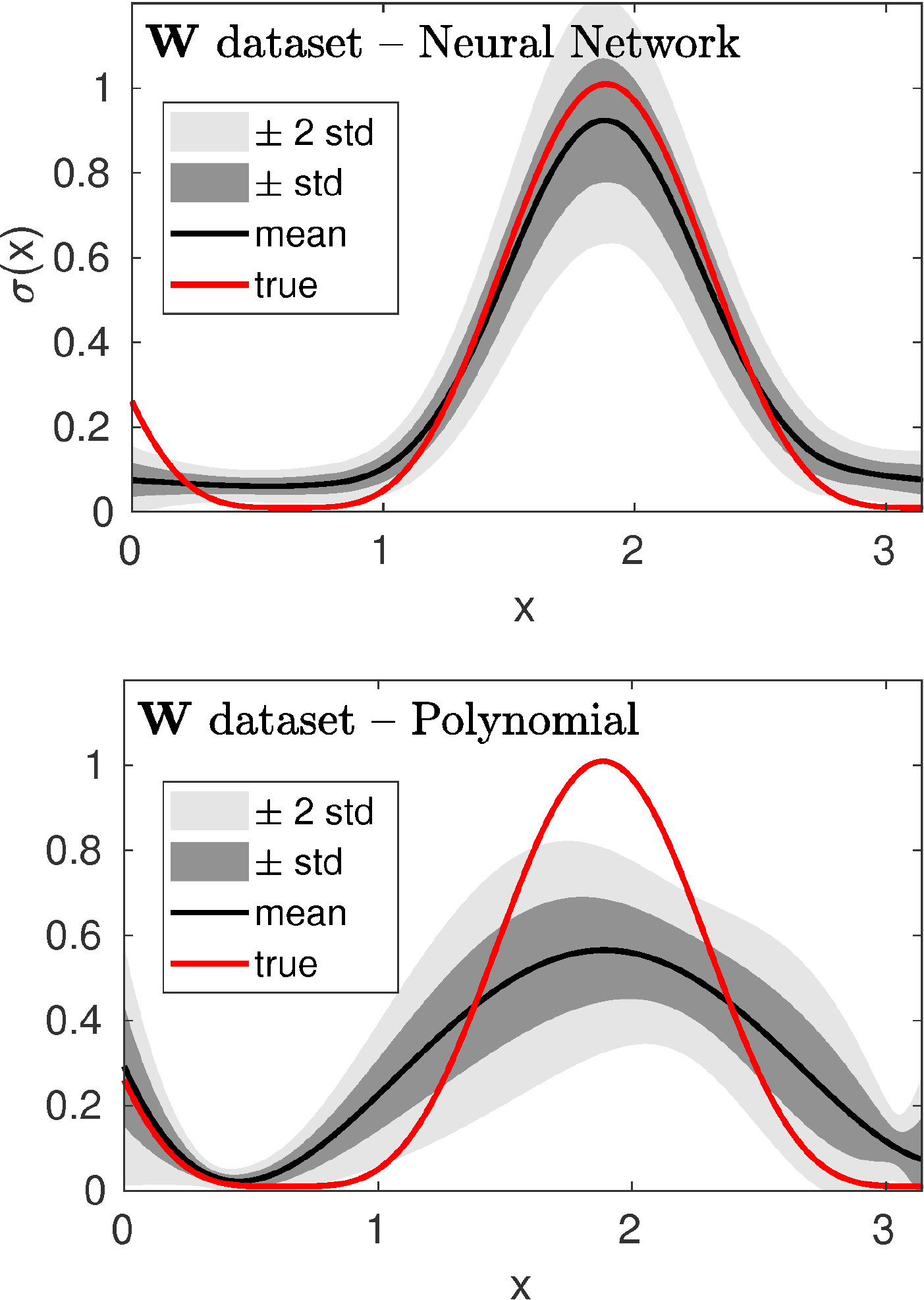}}
\caption{{\bf W} dataset: True value of the standard deviation $\sigma$ (red line) and mean value obtained averaging over 100 independent runs (black line). The gray shaded areas denote the confidence interval of one and two standard deviations calculated from the same ensemble of runs. In the top panel $\sigma$ is calculated through a Neural Network, while in the bottom panel as a polynomial function (see text). }
\label{W_dataset}
\end{center}
\vskip -0.1in
\end{figure}

An important point to notice is that the method will inherently attempt to correct any inaccuracy in $f(\mathbf{x})$ by assigning larger variances.  Fir instance, the agreement between predicted and true values of the standard deviation $\sigma$ presented in Figures \ref{G_dataset}-\ref{W_dataset} must be understood within the limits of the approximation of the mean function (provided by a Gaussian Process regression in those toy examples).

By decoupling the prediction of the mean function from the estimation of the variance, this method is not very expensive, and it is suitable for large datasets. Moreover, for the same reason this method is very appealing in all applications where the mean function is necessarily computed via an expensive black-box, such as computer simulations, for which the de-facto standard of uncertainty quantification is based on running a large (time-consuming and expensive) ensemble, and for which large dataset of archived runs are often avaiable.
Finally, the formulation is well suited for high-dimensional problems, since the cost function is calculated point-wise for any instance of prediction and observation.\\
Although very simple and highly efficient the method is still fully parametric, and hence it bears the usual drawback of possibly dealing with a large number of choices for the model selection. Interesting future directions will be to incorporate the Accuracy-Reliability cost function in a non-parametric Bayesian method for heteroskedastic regression and to generalize the constraint of Gaussian residuals.

\bibliographystyle{plainnat}
\bibliography{icml_bib}

\begin{thebibliography}{34}
\providecommand{\natexlab}[1]{#1}
\providecommand{\url}[1]{\texttt{#1}}
\expandafter\ifx\csname urlstyle\endcsname\relax
  \providecommand{\doi}[1]{doi: #1}\else
  \providecommand{\doi}{doi: \begingroup \urlstyle{rm}\Url}\fi

\bibitem[Anderson(1996)]{anderson96}
Jeffrey~L Anderson.
\newblock A method for producing and evaluating probabilistic forecasts from
  ensemble model integrations.
\newblock \emph{Journal of Climate}, 9\penalty0 (7):\penalty0 1518--1530, 1996.

\bibitem[Br{\"o}cker and Smith(2007)]{brocker07}
Jochen Br{\"o}cker and Leonard~A Smith.
\newblock Scoring probabilistic forecasts: The importance of being proper.
\newblock \emph{Weather and Forecasting}, 22\penalty0 (2):\penalty0 382--388,
  2007.

\bibitem[Gal and Ghahramani(2016)]{gal16}
Yarin Gal and Zoubin Ghahramani.
\newblock Dropout as a bayesian approximation: Representing model uncertainty
  in deep learning.
\newblock In \emph{international conference on machine learning}, pages
  1050--1059, 2016.

\bibitem[Genz(1984)]{genz84}
Alan Genz.
\newblock Testing multidimensional integration routines.
\newblock In \emph{Proc. Of International Conference on Tools, Methods and
  Languages for Scientific and Engineering Computation}, pages 81--94, New
  York, NY, USA, 1984. Elsevier North-Holland, Inc.
\newblock ISBN 0-444-87570-0.

\bibitem[Gneiting and Katzfuss(2014)]{gneiting14}
Tilmann Gneiting and Matthias Katzfuss.
\newblock Probabilistic forecasting.
\newblock \emph{Annual Review of Statistics and Its Application}, 1:\penalty0
  125--151, 2014.

\bibitem[Gneiting et~al.(2005)Gneiting, Raftery, Westveld~III, and
  Goldman]{gneiting05}
Tilmann Gneiting, Adrian~E Raftery, Anton~H Westveld~III, and Tom Goldman.
\newblock Calibrated probabilistic forecasting using ensemble model output
  statistics and minimum crps estimation.
\newblock \emph{Monthly Weather Review}, 133\penalty0 (5):\penalty0 1098--1118,
  2005.

\bibitem[Gneiting et~al.(2007)Gneiting, Balabdaoui, and Raftery]{gneiting2007}
Tilmann Gneiting, Fadoua Balabdaoui, and Adrian~E Raftery.
\newblock Probabilistic forecasts, calibration and sharpness.
\newblock \emph{Journal of the Royal Statistical Society: Series B (Statistical
  Methodology)}, 69\penalty0 (2):\penalty0 243--268, 2007.

\bibitem[Goldberg et~al.(1998)Goldberg, Williams, and Bishop]{goldberg98}
Paul~W Goldberg, Christopher~KI Williams, and Christopher~M Bishop.
\newblock Regression with input-dependent noise: A gaussian process treatment.
\newblock In \emph{Advances in neural information processing systems}, pages
  493--499, 1998.

\bibitem[Guo et~al.(2017)Guo, Pleiss, Sun, and Weinberger]{guo2017}
Chuan Guo, Geoff Pleiss, Yu~Sun, and Kilian~Q Weinberger.
\newblock On calibration of modern neural networks.
\newblock In \emph{Proceedings of the 34th International Conference on Machine
  Learning-Volume 70}, pages 1321--1330. JMLR. org, 2017.

\bibitem[Hamill(1997)]{hamill97}
Thomas~M Hamill.
\newblock Reliability diagrams for multicategory probabilistic forecasts.
\newblock \emph{Weather and forecasting}, 12\penalty0 (4):\penalty0 736--741,
  1997.

\bibitem[Hamill(2001)]{hamill01}
Thomas~M Hamill.
\newblock Interpretation of rank histograms for verifying ensemble forecasts.
\newblock \emph{Monthly Weather Review}, 129\penalty0 (3):\penalty0 550--560,
  2001.

\bibitem[Hern{\'a}ndez-Lobato and Adams(2015)]{hernandez15}
Jos{\'e}~Miguel Hern{\'a}ndez-Lobato and Ryan Adams.
\newblock Probabilistic backpropagation for scalable learning of bayesian
  neural networks.
\newblock In \emph{International Conference on Machine Learning}, pages
  1861--1869, 2015.

\bibitem[Hersbach(2000)]{hersbach00}
Hans Hersbach.
\newblock Decomposition of the continuous ranked probability score for ensemble
  prediction systems.
\newblock \emph{Weather and Forecasting}, 15\penalty0 (5):\penalty0 559--570,
  2000.

\bibitem[Johnson and Bowler(2009)]{johnson09}
Christine Johnson and Neill Bowler.
\newblock On the reliability and calibration of ensemble forecasts.
\newblock \emph{Monthly Weather Review}, 137\penalty0 (5):\penalty0 1717--1720,
  2009.

\bibitem[Kersting et~al.(2007)Kersting, Plagemann, Pfaff, and
  Burgard]{kersting07}
Kristian Kersting, Christian Plagemann, Patrick Pfaff, and Wolfram Burgard.
\newblock Most likely heteroscedastic gaussian process regression.
\newblock In \emph{Proceedings of the 24th international conference on Machine
  learning}, pages 393--400. ACM, 2007.

\bibitem[Kuleshov et~al.(2018)Kuleshov, Fenner, and Ermon]{kuleshov2018}
Volodymyr Kuleshov, Nathan Fenner, and Stefano Ermon.
\newblock Accurate uncertainties for deep learning using calibrated regression.
\newblock \emph{arXiv preprint arXiv:1807.00263}, 2018.

\bibitem[Lakshminarayanan et~al.(2017)Lakshminarayanan, Pritzel, and
  Blundell]{lakshminarayanan17}
Balaji Lakshminarayanan, Alexander Pritzel, and Charles Blundell.
\newblock Simple and scalable predictive uncertainty estimation using deep
  ensembles.
\newblock In \emph{Advances in Neural Information Processing Systems}, pages
  6405--6416, 2017.

\bibitem[Leutbecher and Palmer(2008)]{leutbecher08}
Martin Leutbecher and Tim~N Palmer.
\newblock Ensemble forecasting.
\newblock \emph{Journal of Computational Physics}, 227\penalty0 (7):\penalty0
  3515--3539, 2008.

\bibitem[Levi et~al.(2019)Levi, Gispan, Giladi, and Fetaya]{levi2019}
Dan Levi, Liran Gispan, Niv Giladi, and Ethan Fetaya.
\newblock Evaluating and calibrating uncertainty prediction in regression
  tasks.
\newblock \emph{arXiv preprint arXiv:1905.11659}, 2019.

\bibitem[MacKay(1992)]{mackay92}
David~JC MacKay.
\newblock A practical bayesian framework for backpropagation networks.
\newblock \emph{Neural computation}, 4\penalty0 (3):\penalty0 448--472, 1992.

\bibitem[Matheson and Winkler(1976)]{matheson76}
James~E Matheson and Robert~L Winkler.
\newblock Scoring rules for continuous probability distributions.
\newblock \emph{Management science}, 22\penalty0 (10):\penalty0 1087--1096,
  1976.

\bibitem[Murphy(1977)]{murphy77}
Allan~H Murphy.
\newblock The value of climatological, categorical and probabilistic forecasts
  in the cost-loss ratio situation.
\newblock \emph{Monthly Weather Review}, 105\penalty0 (7):\penalty0 803--816,
  1977.

\bibitem[Murphy and Winkler(1992)]{murphy92}
Allan~H Murphy and Robert~L Winkler.
\newblock Diagnostic verification of probability forecasts.
\newblock \emph{International Journal of Forecasting}, 7\penalty0 (4):\penalty0
  435--455, 1992.

\bibitem[Neal(2012)]{neal12}
Radford~M Neal.
\newblock \emph{Bayesian learning for neural networks}, volume 118.
\newblock Springer Science \& Business Media, 2012.

\bibitem[Owens et~al.(2014)Owens, Horbury, Wicks, McGregor, Savani, and
  Xiong]{owens14}
Mathew~J Owens, TS~Horbury, RT~Wicks, SL~McGregor, NP~Savani, and M~Xiong.
\newblock Ensemble downscaling in coupled solar wind-magnetosphere modeling for
  space weather forecasting.
\newblock \emph{Space Weather}, 12\penalty0 (6):\penalty0 395--405, 2014.

\bibitem[Platt et~al.(1999)]{platt1999}
John Platt et~al.
\newblock Probabilistic outputs for support vector machines and comparisons to
  regularized likelihood methods.
\newblock \emph{Advances in large margin classifiers}, 10\penalty0
  (3):\penalty0 61--74, 1999.

\bibitem[Rasmussen and Williams(2006)]{rasmussen06}
Carl~Edward Rasmussen and Christopher~KI Williams.
\newblock \emph{Gaussian process for machine learning}.
\newblock MIT press, 2006.

\bibitem[Song et~al.(2019)Song, Diethe, Kull, and Flach]{song2019}
Hao Song, Tom Diethe, Meelis Kull, and Peter Flach.
\newblock Distribution calibration for regression.
\newblock \emph{arXiv preprint arXiv:1905.06023}, 2019.

\bibitem[Toth et~al.(2003)Toth, Talagrand, Candille, and Zhu]{toth03}
Zoltan Toth, Olivier Talagrand, Guillem Candille, and Yuejian Zhu.
\newblock Probability and ensemble forecasts, 2003.

\bibitem[Weigend and Nix(1994)]{weigend94}
Andreas~S Weigend and David~A Nix.
\newblock Predictions with confidence intervals (local error bars).
\newblock In \emph{Proceedings of the international conference on neural
  information processing}, pages 847--852, 1994.

\bibitem[Wilks(2011)]{wilks11}
Daniel~S Wilks.
\newblock \emph{Statistical methods in the atmospheric sciences}, volume 100.
\newblock Academic press, 2011.

\bibitem[Williams(1996)]{williams96}
Peter~M Williams.
\newblock Using neural networks to model conditional multivariate densities.
\newblock \emph{Neural Computation}, 8\penalty0 (4):\penalty0 843--854, 1996.

\bibitem[Yuan and Wahba(2004)]{yuan04}
Ming Yuan and Grace Wahba.
\newblock Doubly penalized likelihood estimator in heteroscedastic regression.
\newblock \emph{Statistics \& probability letters}, 69\penalty0 (1):\penalty0
  11--20, 2004.

\bibitem[Zadrozny and Elkan(2001)]{zadrozny2001}
Bianca Zadrozny and Charles Elkan.
\newblock Obtaining calibrated probability estimates from decision trees and
  naive bayesian classifiers.
\newblock In \emph{Icml}, volume~1, pages 609--616. Citeseer, 2001.

\end{thebibliography}

\clearpage
\newpage

\end{document}